\documentclass[letterpaper, 10 pt, conference]{ieeeconf}  

\IEEEoverridecommandlockouts                              

\usepackage{amsmath,amsfonts}
\usepackage{algorithm}
\usepackage{array}
\usepackage[caption=false,font=normalsize,labelfont=sf,textfont=sf]{subfig}
\usepackage{textcomp}
\usepackage{stfloats}
\usepackage{url}
\usepackage{verbatim}
\usepackage{graphicx}
\usepackage{cite}
\usepackage{caption}
\usepackage{pdfsync}
\usepackage{xcolor} 
\usepackage{booktabs}
\usepackage{amssymb}

\usepackage[noend]{algpseudocode}
\usepackage{algorithmicx}
\usepackage{comment}
\usepackage{multirow}
\usepackage{threeparttable}
\usepackage{cleveref}
\usepackage{nicematrix}
\usepackage{epstopdf}
\usepackage{longtable}
\usepackage{bm}
\usepackage[normalem]{ulem}
\usepackage{tablefootnote}
\usepackage{leftindex}

\useunder{\uline}{\ul}{}

\DeclareMathOperator*{\argmin}{arg\,min}

\newcommand*{\re}{\textcolor{black}}

\title{\LARGE \bf
{Autonomous Elemental Characterization Enabled by a Low Cost Robotic Platform Built Upon a Generalized Software Architecture}
}

\author{{Xuan Cao$^{1*}$, Yuxin Wu$^{1}$, Michael L. Whittaker$^{1,2*}$}
\thanks{$^{1}$Energy Geosciences Division, Lawrence Berkeley National Laboratory, Berkeley, CA 94720, United States. $^{2}$Materials Science Division, Lawrence Berkeley National Laboratory, Berkeley, CA 94720, United States. $^{*}$To whom correspondence should be addressed.
Contact:
        {\tt\small caoxuan8872@gmail.com} and {\tt\small mwhittaker@lbl.gov}.}%
}

\begin{document}

\maketitle
\thispagestyle{empty}
\pagestyle{empty}

\begin{abstract}
Despite the rapidly growing applications of robots in industry, the use of robots to automate tasks in scientific laboratories is less prolific due to lack of generalized methodologies and high cost of hardware. 
This paper focuses on the automation of characterization tasks necessary for reducing cost while maintaining generalization, 
and proposes a software architecture for building robotic systems in scientific laboratory environment.
A dual-layer (Socket.IO and ROS) action server design is the basic building block, which facilitates the implementation of a web-based front end for user-friendly operations and the use of ROS Behavior Tree for convenient task planning and execution. 
A robotic platform for automating mineral and material sample characterization is built upon the architecture, with an open source, low-cost three-axis computer numerical control gantry system serving as the main robot.
A handheld laser induced breakdown spectroscopy (LIBS) analyzer is integrated with a 3D printed adapter, enabling automated 2D chemical mapping. We demonstrate the utility of automated chemical mapping by scanning of the surface of a spodumene-bearing pegmatite core
sample with a 1071-point dense hyperspectral map acquired at a rate of 1520 bits per second. 
Automated LIBS scanning enables controlled chemical quantification in the laboratory that complements field-based measurements acquired with the same handheld device, linking resource exploration and processing steps in the supply chain for lithium-based battery materials.

\end{abstract}

\section{Introduction}

The rapid development of robotics in recent years has given a boost of its applications in \textit{industry}, such as machine tending~\cite{jia2024review}, palletizing~\cite{lamon2020towards}, and assembly~\cite{faccio2019collaborative}. The operational stock of industrial robots worldwide increased from about 1.3 million in 2013 to 4.3 million in 2023~\cite{worldrobotics2024}.

Similarly, robotic automation in \textit{research laboratories} has become an emerging field, since ``Robotics and automation can enable scientific experiments to be conducted faster, more safely, more accurately, and with greater reproducibility, allowing scientists to tackle large societal problems in domains such as health and energy on a shorter timescale''~\cite{angelopoulos2024}.
Although there have been successful applications of robotic automation in laboratories~\cite{szymanski2023autonomous, jiang2023autonomous, asche2021robotic, burger2020mobile, dai2024autonomous}, the use of robots to automate laboratory operations is still limited in general due to the automation gap caused by the variety of tasks and protocols~\cite{holland2020automation}, ultimately resulting in high costs.

This work sheds some light on the automation of \textit{characterization} tasks in labs, which determine the properties, composition, and behavior of substances (e.g. spectrometry, microscopy, thermal analysis, etc.), and hence are essential in scientific research. 
One common pattern in characterization tasks is \textit{sample-move-instrument-stay} (SMIS), where a sample is placed to a specific position for an analytical instrument to start working. Automating this pattern using robots requires precise \textit{pick-and-place} operations and enough degrees of freedom.

By contrast, this paper focuses on the \textit{sample-stay-instrument-move} (SSIM) pattern, where an instrument is held by a robot and moved around a sample during characterization. Automating this pattern does not require \textit{pick-and-place} operations since the instrument is mounted on the robot all the time. Sample standardization, such as positioning on a 2-d horizontal plane, reduces the robot's required degrees of freedom to reach the samples, which could potentially lower the hardware cost.

Towards this end, this paper introduces a robotic platform for automating SSIM characterization tasks for mineral and material samples. The platform consists of: (1)~a low-cost 3D (translational movements in X, Y, and Z directions) gantry system commonly used in traditional computer numerical control (CNC) machining as the primary robot, (2)~an analytical instrument mounted to the gantry system for sample characterization, and (3)~a stereo camera capable of depth sensing for locating samples to be measured. All components, and samples to be measured, are placed on a benchtop. The general workflow consists of the following steps: (1)~a sample location is either predefined, or else identified by the camera; (2)~the gantry system takes the analytical instrument to the sample location; (3)~the instrument starts characterization and collects raw data; (4)~the raw data are processed and optional feedback is generated. 


The core of the software is a generalized custom-designed architecture for automation systems in laboratory environments. The basis of the architecture is a \textit{dual-layer action server} design for every hardware component, which monitors incoming operation requests through both \textit{Socket.IO}~\cite{rai2013socket} and \textit{Robot Operating System} (ROS)~\cite{abm6074} communication protocols and commands the hardware to act accordingly.   
On top of all action servers lies a Behavior Tree (BT)~\cite{colledanchise2018behavior} which orchestrates the hardware components by interacting with their action servers to automate the characterization workflow.
A web-based front end is developed to ensure user-friendly operations of the platform, including both manual control of each individual hardware and execution of the BT.

To showcase the efficacy of the platform, we integrate a handheld laser induced breakdown
spectroscopy (LIBS) analyzer to the gantry system and use the platform to perform dense LIBS scanning on the surface of a spodumene-bearing pegmatite core sample with 1071 measurement points, each containing optical emission spectra between 190 nm and 950 nm with 0.03 nm resolution, corresponding to 22800 data channels per measurement. The resulting $2 \times 10^{7}$ data are automatically quantified using a custom algorithm, yielding spatially-resolved, comprehensive chemical analysis with parts-per-million levels for most chemical elements.
The automated LIBS scanning (1)~frees researchers from tedious operations,
(2)~accelerates LIBS characterization by at least 3 times the rate of manual operations, and (3)~provides crucial information about downstream processing chemistry.

This paper makes three contributions. 
First, a generalized software architecture for building robotic automation systems in scientific laboratory environment is proposed. 
Second, a low-cost gantry system commonly used in CNC machining is shown to be capable of working as a robot for the automation of SMIS characterization tasks.  
Third, automated dense LIBS scanning using the developed robotic platform and automatic data reduction is achieved.
 
\section{Related Work}

This section reviews literature related to automation in laboratory environments and LIBS scanning.

Industrial robot arms, installed either on benchtops~\cite{szymanski2023autonomous} or mobile bases~\cite{burger2020mobile, dai2024autonomous}, have been used to automate lab-level experimental protocols, by handling samples, transferring samples between instruments, and operating instruments. Though these applications of robot arms were successful, few technical details were reported in~\cite{szymanski2023autonomous, burger2020mobile, dai2024autonomous} with respect to the automation systems, such as individual instrument control, multi-instrument integration, and high-level planning and execution. 
There have been other automation systems for more narrowed down tasks in laboratory environments with more technical details reported, such as solid dispensing~\cite{jiang2023autonomous}, liquid handling~\cite{fleischer2018analytical}, operating reactors~\cite{asche2021robotic}, simple sample pre-treatments followed by mass spectrometry characterization~\cite{chen2017dual}, and mobile robot navigation in a distributed lab~\cite{liu2013mobile}. However, the approaches introduced in~\cite{jiang2023autonomous, fleischer2018analytical, asche2021robotic, chen2017dual, liu2013mobile} are ad hoc and can be difficult to be generalized. 
In contrast to these studies, this work explores to replace industrial robot arms with a low-cost 3D translation gantry system for SSIM-pattern characterization tasks based on a generalized software architecture for lab automation systems.

This work chooses LIBS~\cite{fabre2020advances} as the characterization probe to validate the robotic platform's competency. 
Plain LIBS measurement is point-wise~\cite{khajehzadeh2015fast} and some extra effort must be made to achieve LIBS scanning.
Some LIBS devices can perform a small step raster pattern within its laser aperture~\cite{connors2016application} but will not suffice for large samples.
Another common approach is putting a sample on a 2D or 3D (X-Y or X-Y-Z) translation stage and moving the sample with respect to the laser beam~\cite{lucena1999mapping, bette2004high, caceres2017megapixel}, which in theory can handle large samples, but falls into the SMIS pattern which can be challenging to automate. By contract, this work presents a SSIM-pattern LIBS scanning which is suitable for large samples and easier to automate with a lower budget.




\begin{table*}[]
\centering
\caption{Desired properties and justification of the designed software architecture.}
\label{tab:sw_property}
\resizebox{\textwidth}{!}{%
\begin{tabular}{l|l|l}
\hline
\textbf{Desired Property} &
  \textbf{Explanation} &
  \textbf{Software Architecture Justification} \\ \hline
Modularity &
  \begin{tabular}[c]{@{}l@{}}The software can be separated into independent \\ modules to enable flexible and reusable implementations.\end{tabular} &
  \begin{tabular}[c]{@{}l@{}}The dual-layer action server blocks, BT and \\ front end  have minimal connnections with each other.\end{tabular} \\ \hline
Adaptability &
  \begin{tabular}[c]{@{}l@{}}The software can be adapted easily \\ for changes in experimental or protocols.\end{tabular} &
  \begin{tabular}[c]{@{}l@{}}The system achieves different experimental \\ protocols by simply executing different BTs.\end{tabular} \\ \hline
Scalability &
  \begin{tabular}[c]{@{}l@{}}It should be convenient to add \\ more devices to the existing system.\end{tabular} &
  \begin{tabular}[c]{@{}l@{}}New dual-layer actoin server blocks can be easily \\ added without changing existing action server blocks.\end{tabular} \\ \hline
Distributed System Support &
  \begin{tabular}[c]{@{}l@{}}The software can be deployed to separate \\ machines since devices can take a high space span.\end{tabular} &
  \begin{tabular}[c]{@{}l@{}}The dual-layer action server blocks, BT and \\ front end can all run in separate machines within \\ the same local area network.\end{tabular} \\ \hline
ROS Integration &
  \begin{tabular}[c]{@{}l@{}}ROS is a widely used set of software libraries and \\ tools for building robotic applications and \\ the software should allow easy integration \\ with ROS for using existing resources.\end{tabular} &
  The ROS layer allows convenient intergration with ROS. \\ \hline
User-friendly Operation &
  \begin{tabular}[c]{@{}l@{}}The software should be easy to learn and use, \\ especially for non-experts of computer science.\end{tabular} &
  The front end allows user-friendly operation. \\ \hline
\end{tabular}%
}
\end{table*}

\section{Generalized Software Architecture} \label{sec:gen_sw}
This section introduces the software architecture designed for the mineral and material sample characterization platform and applicable to laboratory automation systems more broadly. 
First, a general scenario  of laboratory automation problems is described and the task of software architecture design is formalized.
Second, the \textit{dual-layer action server} design for any individual device's communication and control, which serves as a basic module of the software architecture, is introduced.
Third, the complete software architecture is demonstrated.

\subsection{A General Scenario of Laboratory Automation Problems} \label{sec:general_scenario}
Let there be a set of \textit{devices} $\mathcal{D} = \{d_{1},\cdots, d_{N}\}$ operated in a laboratory environment.
A device can be either a scientific instrument (e.g. a furnace, a balance, a spectrometer, etc.) or a robotic tool (e.g. a robot arm, a gripper, a camera, etc.).
Each device $d_{i} \in \mathcal{D} $ can perform a set of \textit{actions}, denoted by $\mathcal{A}_i = \{a_1^i,\cdots,a_{M_i}^i\}$, where $|\mathcal{A}_i| = M_i$ indicates devices may have different numbers of actions. For example, a balance's action set can be $\{\tt{tare, zero, weigh}\}$ and a gripper's action set can be $\{\tt{open, close}\}$.
Let $\mathbb{A} = \bigcup_{i = 1}^{N} \mathcal{A}_i$ be the set of \textit{all actions of all devices}. Then the automation of an experimental protocol $\mathcal{P}$, a sequence of sets of actions, can be denoted by
\begin{equation*}
    \mathcal{P} = [A_1, \cdots, A_T],
\end{equation*}
where $A_t \subseteq \mathbb{A}, |A_t| \geq 1, \forall t \in [1, T]$. In other words, at any step $t$, there can be either one action or multiple actions running. One automation system may need to handle multiple protocols. Let be $\mathbb{P} = \{\mathcal{P}_1,\cdots,\mathcal{P}_K\}$ be the set of protocols handled by one automation system. 

We propose that the software for orchestrating $\mathcal{D}$ to achieve $\mathcal{P}_i \in \mathbb{P}$ should satisfy the following properties:
modularity, adaptability, scalability, distributed system support, ROS integration, and user-friendly operation. Explanations of these properties are listed in Table~\ref{tab:sw_property}.
Next, we introduce the \textit{dual-layer action server} software design as a basic building block and justify the complete software architecture built upon it. 

\subsection{Dual-layer Action Server Design} \label{sec:two_layer_action_server}

\begin{figure}[]
    \centering
    \includegraphics[width=3in]{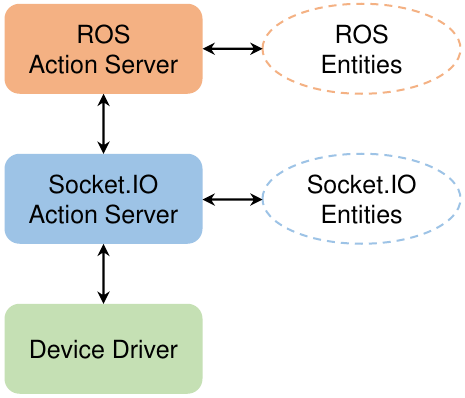}
    \caption{The dual-layer action server design for any individual device's communication and control.
    }
    \label{fig:single_device_control}
\end{figure}

Fig.~\ref{fig:single_device_control} shows the \textit{dual-layer action server} software design for any individual device's communication and control. 
First, upon the device's driver is a Socket.IO \textit{action server} layer. It receives requests from Socket.IO clients and handles those requests by calling the device's driver. 
Second, beyond the Socket.IO layer is a ROS action server layer. It incorporates a corresponding Socket.IO client and serves as a bridge between ROS action clients and the Socket.IO action server.
Communications with the device can flow via either the Socket.IO layer or the ROS layer, whichever is more convenient, ensuring high flexibility and efficiency. Note that it is a subjective design choice to use Socket.IO as the first layer communication protocol for its convenient built-in features like reconnection and broadcasting. It is possible to replace Socket.IO with other communication protocols such as WebSocket~\cite{websocket}. 

\subsection{Complete Software Architecture} \label{sec:complete_software_architecture}

\begin{figure*}[]
    \centering
    \includegraphics[width=7in]{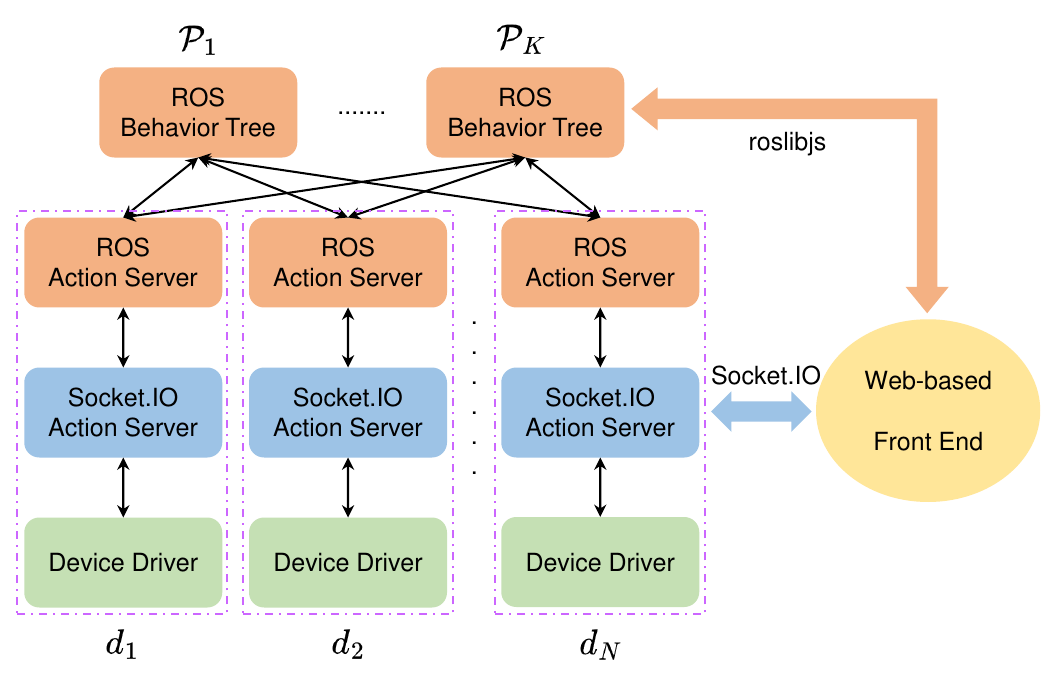}
    \caption{The complete software architecture designed for general laboratory automation systems. $\{d_1,\cdots,d_N\}$ denotes the set of devices integrated into the system and $\{\mathcal{P}_1,\cdots,\mathcal{P}_K\}$ denotes the set of experimental protocols represented by behavior trees to be executed.
    }
    \label{fig:sw_architecture}
\end{figure*}

Fig.~\ref{fig:sw_architecture} shows the complete software architecture for a laboratory automation system. 
Recall the system consists of a set of devices $\mathcal{D} = \{d_1,\cdots,d_N\}$ and handles a set of experimental protocols $\mathbb{P} = \{\mathcal{P}_1,\cdots,\mathcal{P}_K\}$. Then the dual-layer action server block for each device $d_i \in \mathcal{D}$ (pink dashed box) builds up the foundation of the software architecture. 
On top of all action server blocks lies a BT~\cite{colledanchise2018behavior} for planning and executing an experimental protocol $\mathcal{P}_j \in \mathbb{P}$ by  
incorporating ROS action clients corresponding to the ROS action servers and structuring the switching between action sets in $\mathcal{P}_j$.
In other words, the BT orchestrates $\mathcal{D}$ and tells each device $d_i \in \mathcal{D}$ what to do in real time according to $\mathcal{P}_j$ by interacting with the ROS action servers.
In practice, multiple BTs are implemented beforehand to account for various experimental protocols in $\mathbb{P}$ and users can choose the one to execute that fits best their needs. 
The last piece of the architecture is a web-based front end to ensure user-friendly operations of the automation system. The front end has dual responsibilities: it (1)~communicates with the devices via the Socket.IO layer to allow efficient monitoring and manual control of the system and (2)~connects with the ROS layer through \textit{roslibjs}~\cite{roslibjs} to enable convenient execution of the BT which is usually implemented within ROS. The designed software architecture is justified in Table~\ref{tab:sw_property} according to the desired properties proposed in Sec.~\ref{sec:general_scenario}.

\section{Robotic Platform}
This section introduces the technical details of the robotic platform for material chemical characterization built upon the software architecture described in Sec.~\ref{sec:gen_sw}.
\subsection{Hardware}
\begin{figure}[]
    \centering
    \includegraphics[width=3in]{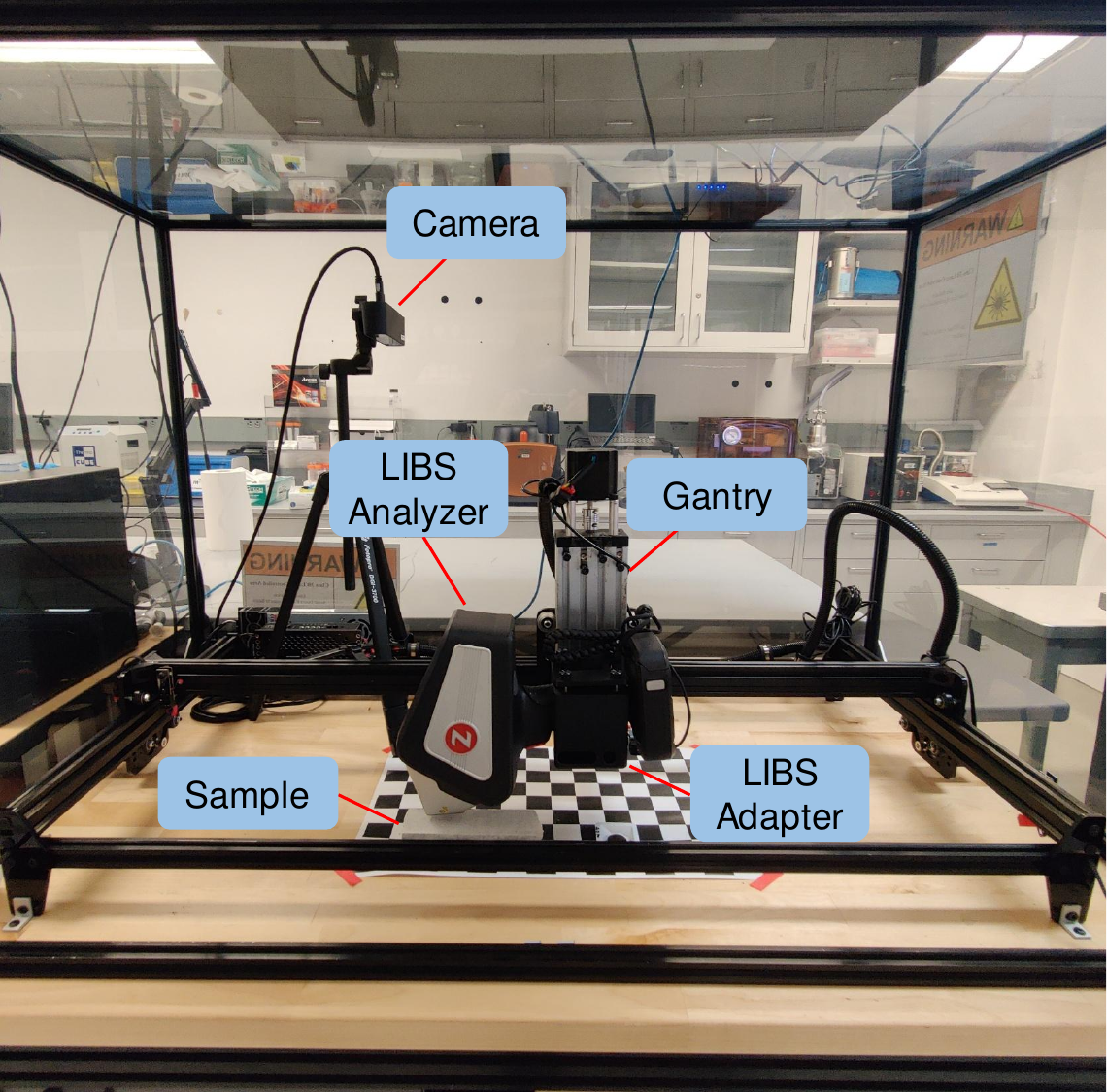}
    \caption{The robotic platform for mineral and material sample characterization tasks.
    }
    \label{fig:components}
\end{figure}

Fig.~\ref{fig:components} shows the hardware components of the robotic platform.
The main body of the platform is a gantry system (LEAD CNC 1010, OpenBuilds) used in CNC machining. It has 3 degrees of freedom, actuating translational movement in X, Y, and Z directions driven by linear stepper motors with 20 $\mu$m resolution and approximately 50 $\mu$m repeatability.
The travel limit (work area) of the gantry is about $730 \times 810\times100 \ (\textrm{mm, X} \times \textrm{Y} \times \textrm{Z})$ and 
is the gantry is fixed on a benchtop within a laser safety enclosure.

An analytical instrument is mounted to the gantry head and approaches samples placed on the work area and perform measurement. 
A handheld LIBS analyzer (Z300, SciAps) is used as the analytical instrument for validating the experiment efficacy of the platform (see details in Sec.~\ref{sec:experiment}). Other characterization probes, such as handheld X-ray fluorescence (XRF) and Raman spectroscopy, can be mounted to the gantry with custom 3D printed adapters. The LIBS analyzer is capable of quantifying the presence of any element on the periodic table, subject to limits of detection that depend on the absolute and relative concentrations, with a spectral range of $190$ to $950$ nm. We use LIBS to show lithium-rich regions of the a mineral sample that will undergo further process for into lithium-ion battery cathode materials.

A stereo depth camera (ZED 2i, StereoLabs) is used to provide the gantry with visual information of samples.  It outputs red, green and blue (RGB) images with a resolution of $1920\times1080$ at a frequency of 30 frames per second (FPS), and depth images with a depth range of $0.3$ to $20$ m at a frequency of up to 100 FPS.


\subsection{Software}
\begin{figure}[]
    \centering
    \includegraphics[width=3in]{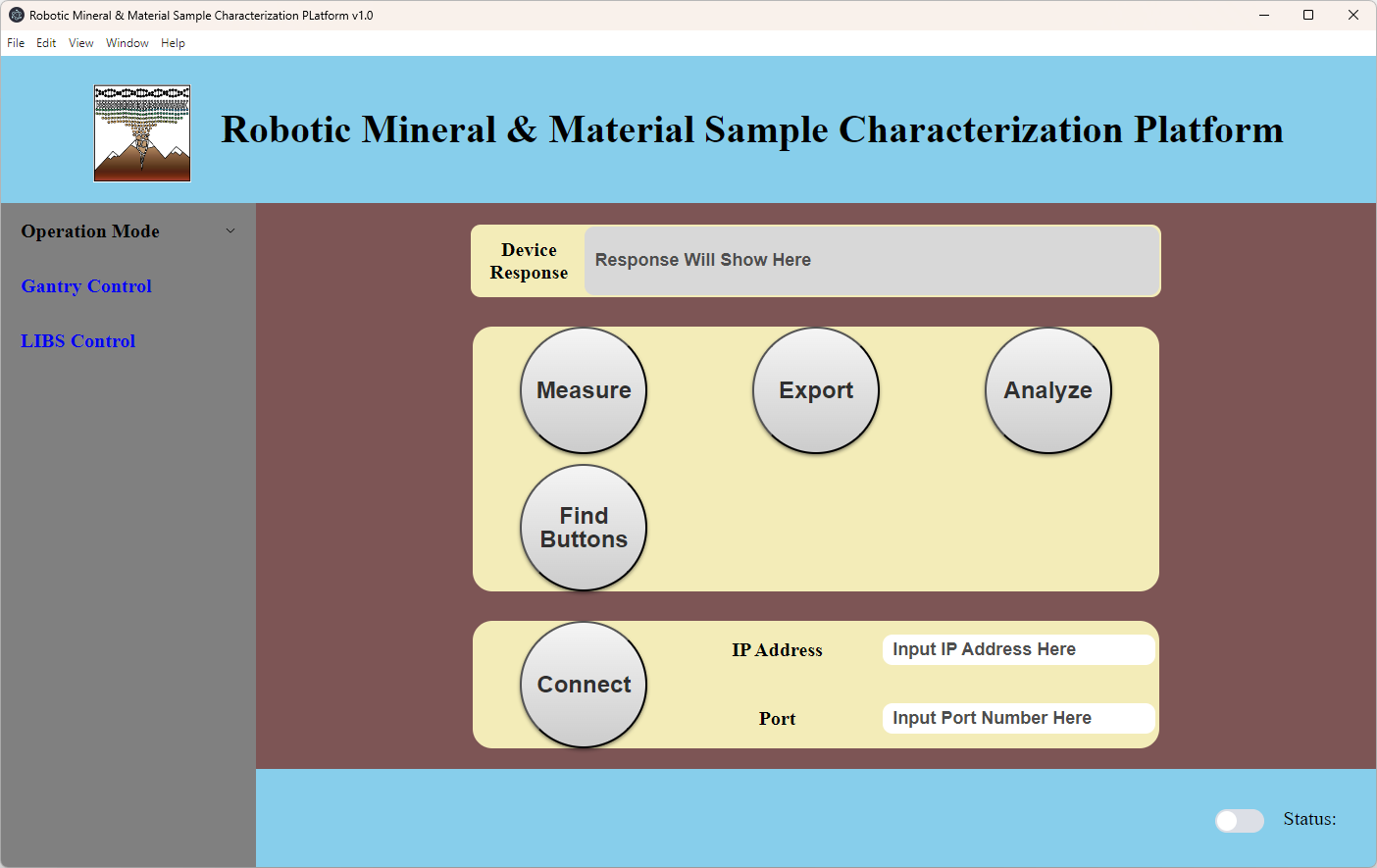}
    \caption{The web-based front end of the robotic platform implemented specially for the experiment described in Sec.~\ref{sec:experiment}
    }
    \label{fig:fe}
\end{figure}

Following the software architecture presented in Sec~\ref{sec:gen_sw} we implement the action servers for the devices, BTs and web-based front end. 
For the gantry, we adopt the existing open source library~\cite{openbuildscontrol} as the Socket.IO action server.
For the LIBS analyzer, due to lack of application programming interfaces (APIs), we additionally implement the device driver based on graphical user interface (GUI) automation, i.e. programmatically controlling the keyboard and mouse to interact with the device's control software.
For the development of the web-based front end, we use \textit{TypeScript}~\cite{bierman2014understanding} and the open source library \textit{Vue3}~\cite{vue3}.
For the implementation of the BTs, we use the open source libraries \textit{py\_trees}~\cite{pytrees} and \textit{py\_trees\_ros}~\cite{pytreesros}.
Fig.~\ref{fig:fe} shows the web-based front end implemented specifically for the experiment described in Sec.~\ref{sec:experiment}.  
For code availability, see Appendix\ref{sec:code}.


\subsection{Vision}
Suppose a measurement can be represented as a point within the stereo camera's view. 
The camera provides both an RGB image and a depth image of the view. The RGB and depth images are aligned with each other, i.e., the RGB and depth information of the measurement point is stored in the same 2D pixel location in the two images, respectively. The point's depth information is the length of the projection of the euclidean distance between the point and the camera onto the camera's Z-axis. 
The point's 2D pixel location and depth information need to be transformed to its 3D location in the world coordinate for the gantry to understand the position of the measurement.  

Formally, let $[u, v]^\intercal$ denote the point's 2D pixel location in the camera's RGB image, $P_c = [x_c,y_c,z_c]^\intercal$ denote that point's 3D location in the camera's coordinate frame $F_c$, and $P_w = [x_w, y_w, z_w]^\intercal$ denote the point's 3D location in the world coordinate frame $F_w$. 
The values of $u$ and $v$ are already known from the camera's RGB image.
The value of $z_c$ is also known from the camera's depth image.
The values of $u, v$ and $z_c$ need to be converted to $P_w$ so that the gantry can carry the LIBS analyzer to the sample location to complete measurement. 

First, the values of $x_c$ and $y_c$ are computed based on the classic pin-hole camera model using the following equations:
\begin{equation*}
  \begin{aligned}
    x_c &= (u - c_x) * z_c / f_x, \\
    y_c &= (v - c_y) * z_c / f_y,
  \end{aligned}
\end{equation*}
where $f_x, f_y, c_x, c_y$ are the camera's intrinsic parameters and provided by the manufacturer. 

Second, $P_c$ is transformed back to $P_w$ with a linear transformation using the following equation:

\begin{equation*}
    P_w = \prescript{c}{w}{R}^{-1} \times (P_c - \prescript{c}{w}{T}),
\end{equation*}
where $\prescript{c}{w}{R}$ and $\prescript{c}{w}{T}$ are the rotational and translational transformation matrices from $F_w$ to $F_c$, and can be computed by the \textit{OpenCV}~\cite{opencv_library} \texttt{solvePnP} function.
The \texttt{solvePnP} function requires a set of points with world coordinates and corresponding image pixel coordinates as inputs. A way to prepare those inputs is to image a black-and-white chess board pattern with known world coordinates of corners, and use the OpenCV \texttt{findChessboardCorners} function to compute the corresponding image pixel coordinates of corners. 






\section{Experiment} \label{sec:experiment}
This section demonstrates the experimental validation of the developed robotic platform. 

\begin{figure}[]
    \centering
    \includegraphics[width=3in]{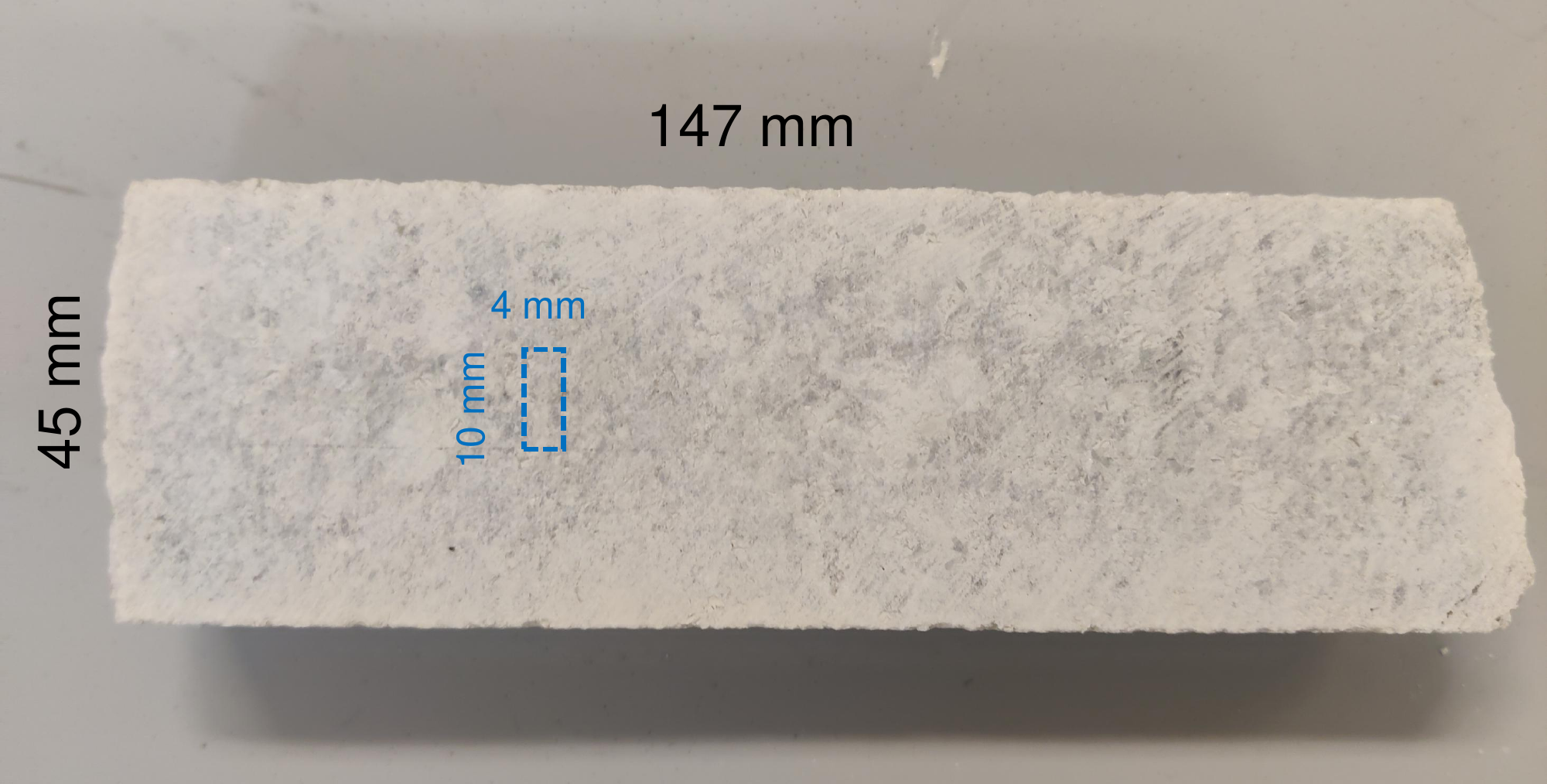}
    \caption{The surface of the spodumene-bearing pegmatite core sample cut by a diamond saw for the LIBS scan experiment. The dashed box indicates the area for the LIBS scan.
    }
    \label{fig:sample}
\end{figure}

\subsection{LIBS Scanning}
We use a LIBS analyzer as the analytical instrument to validate the efficacy of the developed robotic platform. 
LIBS uses pulsed laser to form a high-temperature plasma from a small quantity of  mass ablated from a sample’s surface. Ionized atoms are driven to electronically excited states, which relax to lower energy states upon cooling. Electronic transitions emit photons with characteristic wavelengths that are element-specific and diagnostic of both the type and concentration of the ablated element~\cite{fabre2020advances}.
LIBS is widely applied for elemental measurement of geological samples~\cite{khajehzadeh2015fast} and believed to be ``the optimal way to achieve a first quick screening and then provide valuable data prior to any further laboratory analyses''~\cite{fabre2020advances} for its fast response and wide elemental cover range. 

One drawback of LIBS is that it only measures \textit{one point} at one time and would be inefficient for nonhomogeneous samples.
To address that issue, we demonstrate a dense LIBS \textit{scan of a sample surface} using the developed platform. This is a particularly useful technique for materials containing lithium, such as the spodumene (LiAlSi$_2$O$_6$)-bearing pegmatite core sample shown in \ref{fig:sample}, because LIBS is one of only a few techniques that provide both high fidelity and spatial localization for lithium.

\begin{figure*}[]
    \centering
    \includegraphics[width=6.2in]{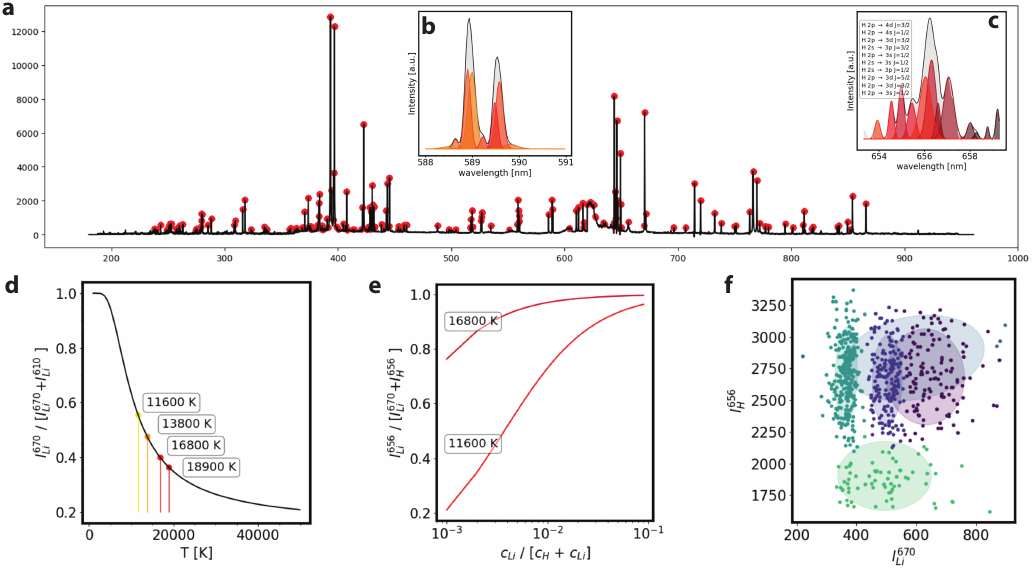}
    \caption{Example of automated data reduction and chemical analysis. (a) Automated peak finding. (b) Na I emission lines. (c) H emission lines. (d) Plasma temperature calibration curves from Li I 670 nm and 610 nm lines. (e) Measured Li concentration fraction as a function of plasma temperature. (f) Four to five mineralogical classes can be identified from Li I and H concentrations using k-means clustering.
    }
    \label{fig:auto_data_reduction}
\end{figure*}

\begin{figure}[]
    \centering
    \includegraphics[width=3in]{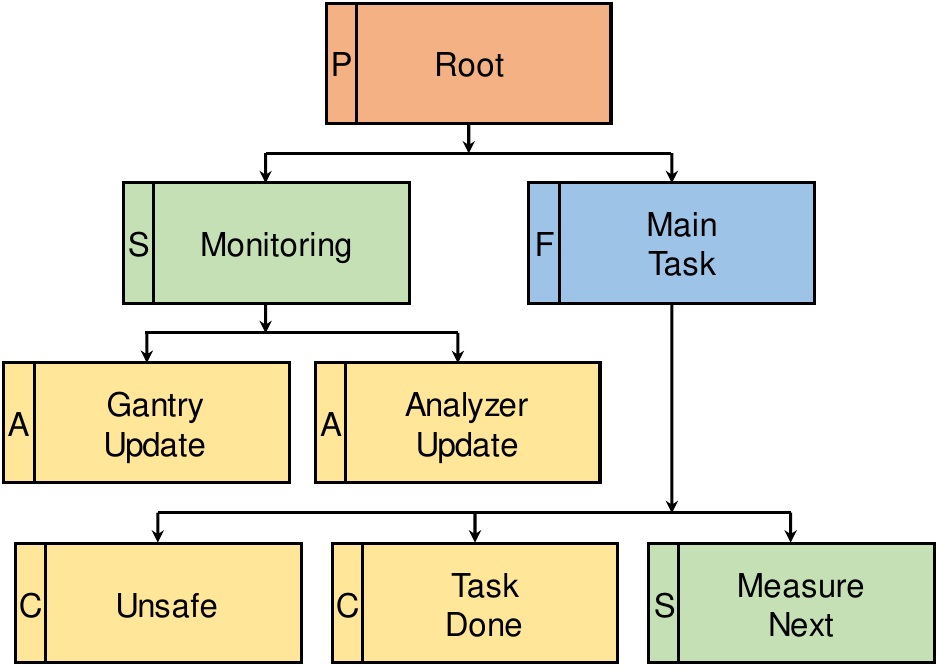}
    \caption{A collapse view of the behavior tree implemented specifically for the experiment described in Sec.~\ref{sec:experiment}. 
    Children of ``Measure Next'' are hidden for simplicity.
    Labels on the left of node blocks indicate node types. P: parallel; S: sequence; F: fallback; A: action; C: condition.
    }
    \label{fig:bt}
\end{figure}

\subsection{Methods}
A cylindrical rock core was cut by a diamond saw to create a flat surface with a dimension of about $147 \times 45$ mm (see Fig.~\ref{fig:sample}) for the experiment. A $4 \times 10$ mm rectangular area on the surface is chosen for a dense LIBS scan with a gap of $0.2$ mm between measurement points in both directions, i.e. there are 1071 LIBS data points collected in total. 

The LIBS analyzer uses a Nd:YAG laser source with a wavelength of 1064 nm and pulse energy of 5 mJ for sample ablation. It is equipped with an on-board spectrometer covering a spectral range of 190 to 950 nm. Before laser ablation argon is flushed to purge the sample surface as well as create an atmosphere that enhances LIBS signals~\cite{palasti2022laser}.


Data is automatically reduced in four steps. First, peaks are found and the background is identified and subtracted from the raw spectrum to produce a data spectrum. Then, peaks are fit with Voigt profiles. Third, the fitted profile is subtracted from the data spectrum and peaks are fit to the residuals to identify interfering peaks. Finally, peaks are indexed to specific elements using an iterative refinement. Fig.~\ref{fig:auto_data_reduction} shows an example of the automated data reduction.

The automated scan process is planned and executed by a specially designed BT (shown in Fig.~\ref{fig:bt}). 
At the beginning, the BT stores the locations of all points to be measured.
For one single point, the gantry first \textit{moves up} to a safe level, then \textit{carries} the LIBS analyzer to location of the point, and finally \textit{moves down} to align the analyzer's aperture with the point.
After that, the analyzer triggers a laser pulse and \textit{collects raw LIBS data}, which is \textit{exported to a csv file} and further \textit{analyzed} by the algorithm described in the previous paragraph.
Then the BT removes that location from the list of all locations. 
The BT ends when all points have been measured. 
Note that the BT is designed in a way that it would pick up any previously failed action before moving forward, ensuring the stability of the scan. 

\begin{figure}[]
    \centering
    \includegraphics[width=3.2in]{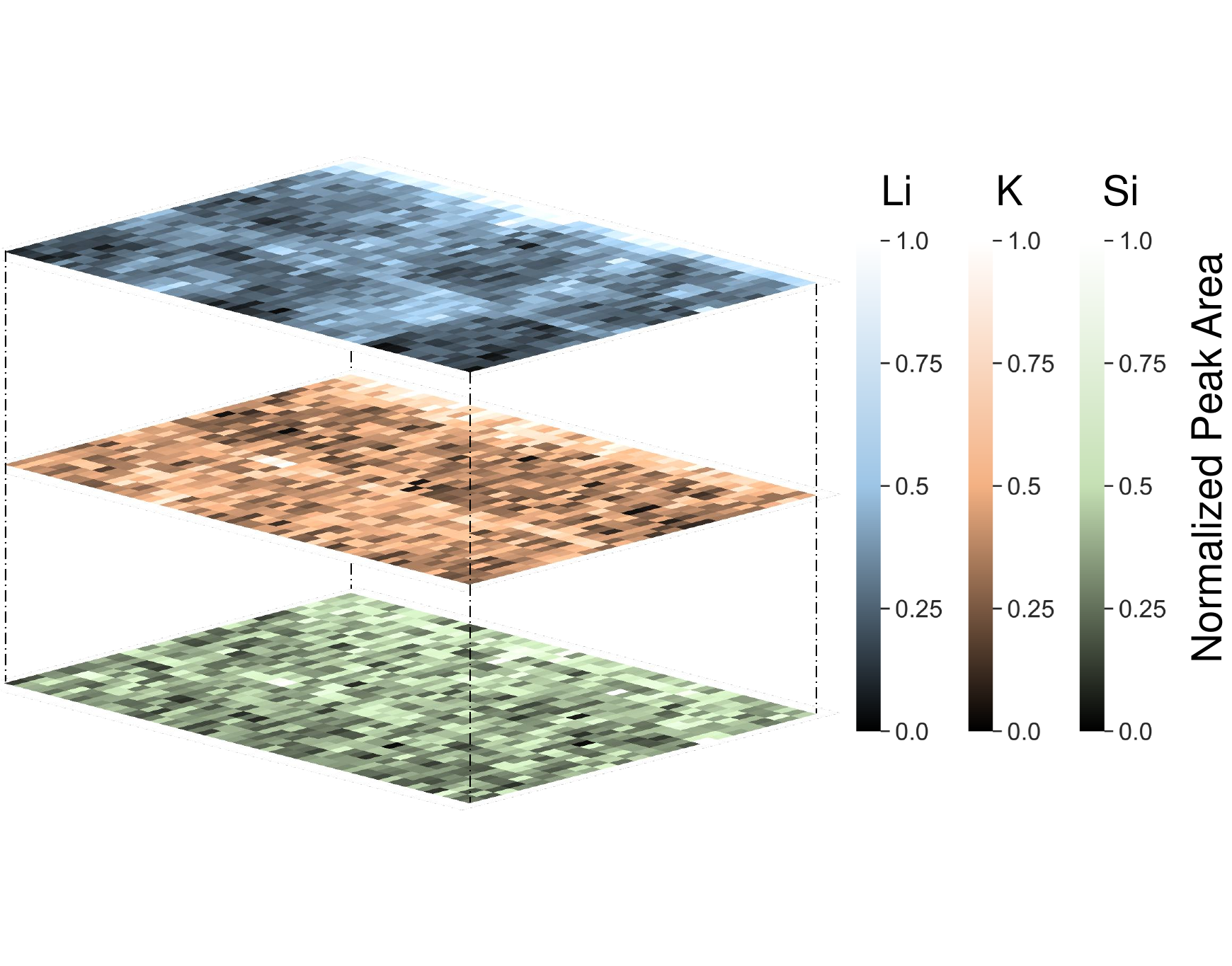}
    \caption{Distributions of various elements (Li, K, and Si) among the scanning area on the sample surface. Elements are characterized by the normalized area of their most predominant peak. Min-max normalization is used to transform the peak areas of each element to $[0, 1]$.
    }
    \label{fig:li_si_mapping}
\end{figure}

\subsection{Result}
Fig.~\ref{fig:li_si_mapping} demonstrates the distributions of various elements (Li, K, and Si) among the scanning area on the sample surface. Each element is characterized by the normalized area of its most predominant peak identified during the automated data reduction. Min-max normalization is used to linearly transform the peak areas of each element to the range of $[0, 1]$. The distributions shown in Fig.~\ref{fig:li_si_mapping} reveals element-specific differences in richness among the scanning area, showcasing the platform's competency in autonomous elemental characterization. 




\section{Discussion}
The average measurement speed of the automated LIBS scan is 1520 bits per second (4 measurements per minute), while a proficient operator can measure one point per minute, indicating the automated LIBS scan not only frees researchers from tedious operations but also accelerates LIBS data acquisition by 3-4 times. 

The primary hardware components, including the gantry, camera, and computer, cost about $\$3000$, $\$550$, and $\$1100$, respectively, which is much more affordable than a commonly used robot arm alone (e.g., a UR5e robot arm manufactured by Universal Robots costs around $\$40000$).
However, the tradeoff is that a commercial robot arm usually has at least 6 degrees of freedom (3D translational movements + 3D rotational movements) and can handle a wider range of automation tasks. 
Even so, the gantry-based approach would be a better choice for tasks with lower requirements of degrees of freedom and prototyping new automation systems with lower budgets.  

The function of the robotic platform is not limited to LISB characterization, rather depends on the tool(s) that can be integrated to the platform. It is possible, and future work should explore, integrating other analytical instruments like a handheld XRF analyzer and tools like an electronically controlled pipette with the platform to cover a wider range of laboratory tasks.

The software architecture is derived from a general scenario of laboratory automation problems and should work well for other laboratory systems in additional to the developed platform. Future work should further validate the generalizability of the software architecture by developing other robotic systems or expanding the current platform. 





\section{Summary and Future Work}
This work introduces a generalized custom-designed software architecture for building robotic automation systems in scientific laboratory environment. The basic block of the architecture is a dual-layer (Socket.IO and ROS) action server design which facilitates the implementation of a web-based front end for user-friendly operations and the use of ROS BT for convenient task planning and execution. Based on the architecture an automated experimental platform is developed, with a low cost gantry system capable of 3D translational movements serving as the main robot. With the integration of a LIBS analyzer, a LIBS scan experiment is conducted to demonstrate the efficacy of the robotic platform and the software architecture.

Future work should add other or more analytical instruments, like an XRF analyzer, to the robotic platform to perform more types of characterization tasks. Future work should also combine other types of tools, like an electronically controlled pipette, with the platform. Another interesting direction for future work is integrating the platform to larger laboratory automation systems to complete more complex experimental protocols.

\addtolength{\textheight}{-1cm}   




\section*{APPENDIX}
\appendices
\subsection{Code Availability} \label{sec:code}
The source code of the LIBS analyzer's driver and Socket.IO server is available at \url{https://github.com/Living-Minerals-Lab/LIBS_trigger}.
The source code of the front end is available at \url{https://github.com/Living-Minerals-Lab/mini_platform_fe}.
The source code of all ROS action servers and BTs is available at \url{https://github.com/Living-Minerals-Lab/mini_platform}. The source code of the automated data reduction is available at \url{https://github.com/Living-Minerals-Lab/LIBZ}.

\section*{Conflicts of interest}
There are no conflicts of interest to declare.

\section*{ACKNOWLEDGMENT}
XC and MLW were supported by the U.S. Department of Energy, Office of Fossil Energy and Carbon Management, Mineral Sustainability Division, through LBNL under Contract DE-AC02-05CH11231.

The authors thank William Evans for designing and 3D printing the LIBS analyzer adapter and assembling the robotic platform.

\bibliographystyle{IEEEtran}
\bibliography{ref.bib}

\begin{thebibliography}{10}
\providecommand{\url}[1]{#1}
\csname url@rmstyle\endcsname
\providecommand{\newblock}{\relax}
\providecommand{\bibinfo}[2]{#2}
\providecommand\BIBentrySTDinterwordspacing{\spaceskip=0pt\relax}
\providecommand\BIBentryALTinterwordstretchfactor{4}
\providecommand\BIBentryALTinterwordspacing{\spaceskip=\fontdimen2\font plus
\BIBentryALTinterwordstretchfactor\fontdimen3\font minus \fontdimen4\font\relax}
\providecommand\BIBforeignlanguage[2]{{%
\expandafter\ifx\csname l@#1\endcsname\relax
\typeout{** WARNING: IEEEtran.bst: No hyphenation pattern has been}%
\typeout{** loaded for the language `#1'. Using the pattern for}%
\typeout{** the default language instead.}%
\else
\language=\csname l@#1\endcsname
\fi
#2}}

\bibitem{jia2024review}
F.~Jia, Y.~Ma, and R.~Ahmad, ``Review of current vision-based robotic machine-tending applications,'' \emph{The International Journal of Advanced Manufacturing Technology}, vol. 131, no.~3, pp. 1039--1057, 2024.

\bibitem{lamon2020towards}
E.~Lamon, M.~Leonori, W.~Kim, and A.~Ajoudani, ``Towards an intelligent collaborative robotic system for mixed case palletizing,'' in \emph{2020 IEEE International Conference on Robotics and Automation (ICRA)}.\hskip 1em plus 0.5em minus 0.4em\relax IEEE, 2020, pp. 9128--9134.

\bibitem{faccio2019collaborative}
M.~Faccio, M.~Bottin, and G.~Rosati, ``Collaborative and traditional robotic assembly: a comparison model,'' \emph{The International Journal of Advanced Manufacturing Technology}, vol. 102, pp. 1355--1372, 2019.

\bibitem{worldrobotics2024}
C.~Müller, ``World robotics 2024 – industrial robots,'' International Federation of Robotics, Tech. Rep., 2024.

\bibitem{angelopoulos2024}
A.~Angelopoulos, J.~F. Cahoon, and R.~Alterovitz, ``Transforming science labs into automated factories of discovery,'' \emph{Science Robotics}, vol.~9, no.~95, p. eadm6991, 2024.

\bibitem{szymanski2023autonomous}
N.~J. Szymanski, B.~Rendy, Y.~Fei, R.~E. Kumar, T.~He, D.~Milsted, M.~J. McDermott, M.~Gallant, E.~D. Cubuk, A.~Merchant, \emph{et~al.}, ``An autonomous laboratory for the accelerated synthesis of novel materials,'' \emph{Nature}, vol. 624, no. 7990, pp. 86--91, 2023.

\bibitem{jiang2023autonomous}
Y.~Jiang, H.~Fakhruldeen, G.~Pizzuto, L.~Longley, A.~He, T.~Dai, R.~Clowes, N.~Rankin, and A.~I. Cooper, ``Autonomous biomimetic solid dispensing using a dual-arm robotic manipulator,'' \emph{Digital Discovery}, vol.~2, no.~6, pp. 1733--1744, 2023.

\bibitem{asche2021robotic}
S.~Asche, G.~J. Cooper, G.~Keenan, C.~Mathis, and L.~Cronin, ``A robotic prebiotic chemist probes long term reactions of complexifying mixtures,'' \emph{Nature Communications}, vol.~12, no.~1, p. 3547, 2021.

\bibitem{burger2020mobile}
B.~Burger, P.~M. Maffettone, V.~V. Gusev, C.~M. Aitchison, Y.~Bai, X.~Wang, X.~Li, B.~M. Alston, B.~Li, R.~Clowes, \emph{et~al.}, ``A mobile robotic chemist,'' \emph{Nature}, vol. 583, no. 7815, pp. 237--241, 2020.

\bibitem{dai2024autonomous}
T.~Dai, S.~Vijayakrishnan, F.~T. Szczypi{\'n}ski, J.-F. Ayme, E.~Simaei, T.~Fellowes, R.~Clowes, L.~Kotopanov, C.~E. Shields, Z.~Zhou, \emph{et~al.}, ``Autonomous mobile robots for exploratory synthetic chemistry,'' \emph{Nature}, pp. 1--8, 2024.

\bibitem{holland2020automation}
I.~Holland and J.~A. Davies, ``Automation in the life science research laboratory,'' \emph{Frontiers in bioengineering and biotechnology}, vol.~8, 2020.

\bibitem{rai2013socket}
R.~Rai, \emph{Socket.IO real-time web application development}.\hskip 1em plus 0.5em minus 0.4em\relax Packt Publishing, 2013.

\bibitem{abm6074}
\BIBentryALTinterwordspacing
S.~Macenski, T.~Foote, B.~Gerkey, C.~Lalancette, and W.~Woodall, ``Robot operating system 2: Design, architecture, and uses in the wild,'' \emph{Science Robotics}, vol.~7, no.~66, p. eabm6074, 2022. [Online]. Available: \url{https://www.science.org/doi/abs/10.1126/scirobotics.abm6074}
\BIBentrySTDinterwordspacing

\bibitem{colledanchise2018behavior}
M.~Colledanchise and P.~{\"O}gren, \emph{Behavior trees in robotics and AI: An introduction}.\hskip 1em plus 0.5em minus 0.4em\relax CRC Press, 2018.

\bibitem{fleischer2018analytical}
H.~Fleischer, D.~Baumann, S.~Joshi, X.~Chu, T.~Roddelkopf, M.~Klos, and K.~Thurow, ``Analytical measurements and efficient process generation using a dual--arm robot equipped with electronic pipettes,'' \emph{Energies}, vol.~11, no.~10, p. 2567, 2018.

\bibitem{chen2017dual}
C.-L. Chen, T.-R. Chen, S.-H. Chiu, and P.~L. Urban, ``Dual robotic arm “production line” mass spectrometry assay guided by multiple arduino-type microcontrollers,'' \emph{Sensors and Actuators B: Chemical}, vol. 239, pp. 608--616, 2017.

\bibitem{liu2013mobile}
H.~Liu, N.~Stoll, S.~Junginger, and K.~Thurow, ``Mobile robot for life science automation,'' \emph{International Journal of Advanced Robotic Systems}, vol.~10, no.~7, p. 288, 2013.

\bibitem{fabre2020advances}
C.~Fabre, ``Advances in laser-induced breakdown spectroscopy analysis for geology: A critical review,'' \emph{Spectrochimica Acta Part B: Atomic Spectroscopy}, vol. 166, p. 105799, 2020.

\bibitem{khajehzadeh2015fast}
N.~Khajehzadeh and T.~K. Kauppinen, ``Fast mineral identification using elemental libs technique,'' \emph{IFAC-PapersOnLine}, vol.~48, no.~17, pp. 119--124, 2015.

\bibitem{connors2016application}
B.~Connors, A.~Somers, and D.~Day, ``Application of handheld laser-induced breakdown spectroscopy (libs) to geochemical analysis,'' \emph{Applied Spectroscopy}, vol.~70, no.~5, pp. 810--815, 2016.

\bibitem{lucena1999mapping}
P.~Lucena, J.~M. Vadillo, and J.~J. Laserna, ``Mapping of platinum group metals in automotive exhaust three-way catalysts using laser-induced breakdown spectrometry,'' \emph{Analytical chemistry}, vol.~71, no.~19, pp. 4385--4391, 1999.

\bibitem{bette2004high}
H.~Bette and R.~Noll, ``High speed laser-induced breakdown spectrometry for scanning microanalysis,'' \emph{Journal of Physics D: Applied Physics}, vol.~37, no.~8, p. 1281, 2004.

\bibitem{caceres2017megapixel}
J.~O. C{\'a}ceres, F.~Pelascini, V.~Motto-Ros, S.~Moncayo, F.~Trichard, G.~Panczer, A.~Mar{\'\i}n-Rold{\'a}n, J.~Cruz, I.~Coronado, and J.~Martin-Chivelet, ``Megapixel multi-elemental imaging by laser-induced breakdown spectroscopy, a technology with considerable potential for paleoclimate studies,'' \emph{Scientific reports}, vol.~7, no.~1, p. 5080, 2017.

\bibitem{websocket}
I.~Fette and A.~Melnikov, ``Rfc 6455: The websocket protocol,'' USA, 2011.

\bibitem{roslibjs}
\BIBentryALTinterwordspacing
RobotWebTools, ``roslibjs,'' Dec. 2023. [Online]. Available: \url{https://github.com/RobotWebTools/roslibjs}
\BIBentrySTDinterwordspacing

\bibitem{openbuildscontrol}
\BIBentryALTinterwordspacing
OpenBuilds, ``Openbuilds-control,'' Dec. 2024. [Online]. Available: \url{https://github.com/OpenBuilds/OpenBuilds-CONTROL}
\BIBentrySTDinterwordspacing

\bibitem{bierman2014understanding}
G.~Bierman, M.~Abadi, and M.~Torgersen, ``Understanding typescript,'' in \emph{ECOOP 2014--Object-Oriented Programming: 28th European Conference, Uppsala, Sweden, July 28--August 1, 2014. Proceedings 28}.\hskip 1em plus 0.5em minus 0.4em\relax Springer, 2014, pp. 257--281.

\bibitem{vue3}
\BIBentryALTinterwordspacing
vuejs, ``core,'' Nov. 2024. [Online]. Available: \url{https://github.com/vuejs/core}
\BIBentrySTDinterwordspacing

\bibitem{pytrees}
\BIBentryALTinterwordspacing
splintered reality, ``py\_trees,'' Jan. 2025. [Online]. Available: \url{https://github.com/splintered-reality/py_trees}
\BIBentrySTDinterwordspacing

\bibitem{pytreesros}
\BIBentryALTinterwordspacing
------, ``py\_trees\_ros,'' Jan. 2025. [Online]. Available: \url{https://github.com/splintered-reality/py_trees_ros}
\BIBentrySTDinterwordspacing

\bibitem{opencv_library}
G.~Bradski, ``{The OpenCV Library},'' \emph{Dr. Dobb's Journal of Software Tools}, 2000.

\bibitem{palasti2022laser}
D.~J. Pal{\'a}sti, L.~P. Villy, A.~Kohut, T.~Ajtai, Z.~Geretovszky, and G.~Galb{\'a}cs, ``Laser-induced breakdown spectroscopy signal enhancement effect for argon caused by the presence of gold nanoparticles,'' \emph{Spectrochimica Acta Part B: Atomic Spectroscopy}, vol. 193, p. 106435, 2022.

\end{thebibliography}

\end{document}